\documentclass[journal]{IEEEtran}
\pdfoutput=1

\RequirePackage[dvipsnames]{xcolor}
\definecolor{myRed}{cmyk}{0,30,20,0}
\definecolor{myBlue}{cmyk}{30,10,0,0}
\definecolor{myPurple}{cmyk}{7,45,0,0}
\definecolor{calvinColor}{cmyk}{0,45,45,0}

\definecolor{myGreen}{cmyk}{100,0,100,0}

\usepackage{colortbl}
\definecolor{manual}{HTML}{e41a1c}
\definecolor{atitControl}{HTML}{ffff33}
\definecolor{Position}{HTML}{4daf4a}
\definecolor{Mission}{HTML}{984ea3}
\definecolor{Stablized}{HTML}{377eb8}

\definecolor{blue1}{HTML}{08519c}
\definecolor{blue2}{HTML}{3182bd}
\definecolor{blue3}{HTML}{6baed6}
\definecolor{blue4}{HTML}{bdd7e7}

\definecolor{green1}{HTML}{31a354}
\definecolor{green2}{HTML}{74c476}
\definecolor{green3}{HTML}{bae4b3}
\definecolor{green4}{HTML}{edf8e9}

\definecolor{purp1}{HTML}{54278f}
\definecolor{purp2}{HTML}{756bb1}
\definecolor{purp3}{HTML}{9e9ac8}
\definecolor{purp4}{HTML}{cbc9e2}

\usepackage{xr}
\usepackage[pagebackref,bookmarks]{hyperref}
\usepackage{cleveref}

\RequirePackage{scalerel}
\RequirePackage{tikz}

\usetikzlibrary{svg.path}

\definecolor{orcidlogocol}{HTML}{A6CE39}
\tikzset{
  orcidlogo/.pic={
    \fill[orcidlogocol] svg{M256,128c0,70.7-57.3,128-128,128C57.3,256,0,198.7,0,128C0,57.3,57.3,0,128,0C198.7,0,256,57.3,256,128z};
    \fill[white] svg{M86.3,186.2H70.9V79.1h15.4v48.4V186.2z}
                 svg{M108.9,79.1h41.6c39.6,0,57,28.3,57,53.6c0,27.5-21.5,53.6-56.8,53.6h-41.8V79.1z M124.3,172.4h24.5c34.9,0,42.9-26.5,42.9-39.7c0-21.5-13.7-39.7-43.7-39.7h-23.7V172.4z}
                 svg{M88.7,56.8c0,5.5-4.5,10.1-10.1,10.1c-5.6,0-10.1-4.6-10.1-10.1c0-5.6,4.5-10.1,10.1-10.1C84.2,46.7,88.7,51.3,88.7,56.8z};
  }
}

\DeclareRobustCommand\orcidiconlinkhelper[2]{%
    \href{https://orcid.org/#2}{#1 \mbox{\scalerel*{%
        \begin{tikzpicture}[yscale=-1,transform shape]%
            \pic{orcidlogo};%
        \end{tikzpicture}%
    }{|}}}%
}

\usepackage[T1]{fontenc}

\usepackage[all]{hypcap}

\hypersetup{
    pdfpagemode=UseNone,
    pdftitle={Predicting UAV Type: An Exploration of Sampling and Data Augmentation for Time Series Classification},
    pdfauthor={Tarik Crnovrsanin, Calvin Yu, Dane Hankamer, Cody Dunne},
    pdfsubject={},
    pdfkeywords={},
    pageanchor=true,
    plainpages=false,       
    hypertexnames=false,    
    bookmarksnumbered=false,
    bookmarksopen=true,     
    bookmarksopenlevel=3,   
    pdfpagemode=UseNone,    
    pdfstartview=Fit,       
    pdfborder={0 0 0},      
    breaklinks=true,        
    colorlinks=true,        
    linkcolor=NavyBlue,
    urlcolor=NavyBlue,
    citecolor=NavyBlue,
    anchorcolor=black,      
    nesting=true,
    linktocpage}
\usepackage{cleveref}
\usepackage{babel}
\usepackage{hhline}
\usepackage{booktabs, multirow} 
\usepackage{soul}
\usepackage{changepage,threeparttable} 
\usepackage{atbegshi}
\usepackage{silence}
\usepackage{caption}
\usepackage{subcaption}
\usepackage{graphicx}
\ifCLASSINFOpdf
\else
\fi
\usepackage{url}

\begin{document}
%
\title{Predicting UAV Type: An Exploration of Sampling and Data Augmentation for Time Series Classification}
%
%

\author{
        \orcidiconlinkhelper{Tarik Crnovrsanin}{0000-0002-4397-5532},
        \orcidiconlinkhelper{Calvin Yu}{0009-0001-5312-7027},
        Dane Hankamer, and
        \orcidiconlinkhelper{Cody Dunne}{0000-0002-1609-9776}
\thanks{Tarik Crnovrsanin, Calvin Yu, and Cody Dunne are with Northeastern University. Emails: [ t.crnovrsanin | yu.calv | c.dunne ]@northeastern.edu.}%
\thanks{Dane Hankamer is with the U.S.\ National Reconnaissance Office and U.S.\ Space Force. Email: hankamda@nro.mil.}%
}

%
%

\markboth{Under submission to IEEE TOC}%
{Shell \MakeLowercase{\textit{et al.}}: Bare Demo of IEEEtran.cls for IEEE Journals}
%



\maketitle

\begin{abstract}
Unmanned aerial vehicles are becoming common and have many productive uses.
However, their increased prevalence raises safety concerns---how can we protect restricted airspace?
Knowing the type of unmanned aerial vehicle can go a long way in determining any potential risks it carries.
For instance, fixed-wing craft can carry more weight over longer distances, thus potentially posing a more significant threat.
This paper presents a machine learning model for classifying unmanned aerial vehicles as quadrotor, hexarotor, or fixed-wing. 
Our approach effectively applies a Long-Short Term Memory (LSTM) neural network for the purpose of time series classification. We performed experiments to test the effects of changing the timestamp sampling method and addressing the imbalance in the class distribution.
Through these experiments, we identified the top-performing sampling and class imbalance fixing methods.
Averaging the macro f-scores across 10 folds of data, we found that the majority quadrotor class was predicted well (98.16\%), and, despite an extreme class imbalance, the model could also predicted a majority of fixed-wing flights correctly (73.15\%). 
Hexarotor instances were often misclassified as quadrotors due to the similarity of multirotors in general (42.15\%).
However, results remained relatively stable across certain methods, which prompted us to analyze and report on their tradeoffs.  
The supplemental material for this paper, including the code and data for running all the experiments and generating the results tables, is available at \url{https://osf.io/mnsgk/}.

\end{abstract}


%
\IEEEpeerreviewmaketitle

\renewcommand{\tabcolsep}{2.5pt}

\section{Introduction}
Unmanned Aerial Vehicles (UAVs) have been increasing in popularity recently due to their compact size, low weight, ease of use, and good maneuverability.
Beyond their military applications, UAVs are also widely used by civilians in support of construction~\cite{construction}, agriculture~\cite{agriculture}, cinema~\cite{film}, and conservation~\cite{conservation}.
These beneficial applications are counterbalanced by the opportunities UAVs create for malicious actions.
For instance, UAVs can be used with low risk to enter areas like airports, prisons, and government buildings \cite{UAV-Contraband, UAV-TSATest, UAV-AntiDrone}.
In the cases of prisons, UAVs can be used to bring in contraband such as phones or drugs.
As the use and reliance on the UAV increases, their potential for misuse also grows.

For an outside observer, it is difficult to tell whether a UAV is behaving properly, maliciously, or is malfunctioning.
Machine learning can aid in this task.
Most current research has focused on identifying anomalous UAV behavior \cite{thresholding, deeplearning, svm}.
As long as the training data shows all anomalous behaviors of interest, these approaches work well.
But a key aspect missed by existing machine learning models is determining the \textit{potential threat} a UAV poses.

To perform a threat assessment of a UAV we must first identify the \textit{characteristics} of the airframe.
We can then use these characteristics to predict the UAV's \textit{capabilities}.
For instance, a fixed-wing UAV can carry more and travel farther.
Whether this payload is explosives, signals intelligence equipment, or illegal drugs, more weight can indicate a larger threat.
Identifying the \textit{type of UAV} has several benefits over detecting anomalies alone. 
First, type detection allows learning from more UAV flight data sets, not just ones with identified anomalous behaviors.
Training with multiple readily-available datasets, possibly in combination, can help us improve the utility of the model.
Second, knowing the type of UAV can assist with other forms of machine learning prediction, such as using UAV flight characteristics to predict its future positions.

In this paper, we use a Long Short-Term Memory (LSTM) model to \textit{predict the type of UAV} flown and explore how different approaches to \textit{preprocessing} and \textit{sampling} the time series affect model performance. 
The LSTM model is particularly advantageous for UAV type classification.
It allows us to input time series data and make classification predictions based on the entire sequence through the use of feedforward and feedback loops.
As making good predictions requires considerable data, we utilize one of the largest public repositories of UAV flight logs---PX4 Flight Review---to establish a prediction baseline.
This dataset poses its own challenges for machine learning.
It was collected over ten years and the UAVs used and collection process changed substantially during that period. 
Moreover, the data contains a serious imbalance in the number of UAVs for each type. 
Later in the paper we will describe how we addressed these challenges and trained an effective LSTM model for UAV type classification.
To the best of our knowledge, our work is the first to predict UAV type from flight log information.

This paper makes the following contributions:
\begin{enumerate}
    \item An LSTM predictive model which effectively classifies the type of UAV flown in a general-purpose flight log,
    \item An exploration of the sampling methods available for time-series trajectory-based classification, and
    \item The results of an experiment on how varied sampling, configuration, and data augmentation approaches affect model performance.
\end{enumerate}

\begin{figure}[t]
\includegraphics[width=.95\linewidth]{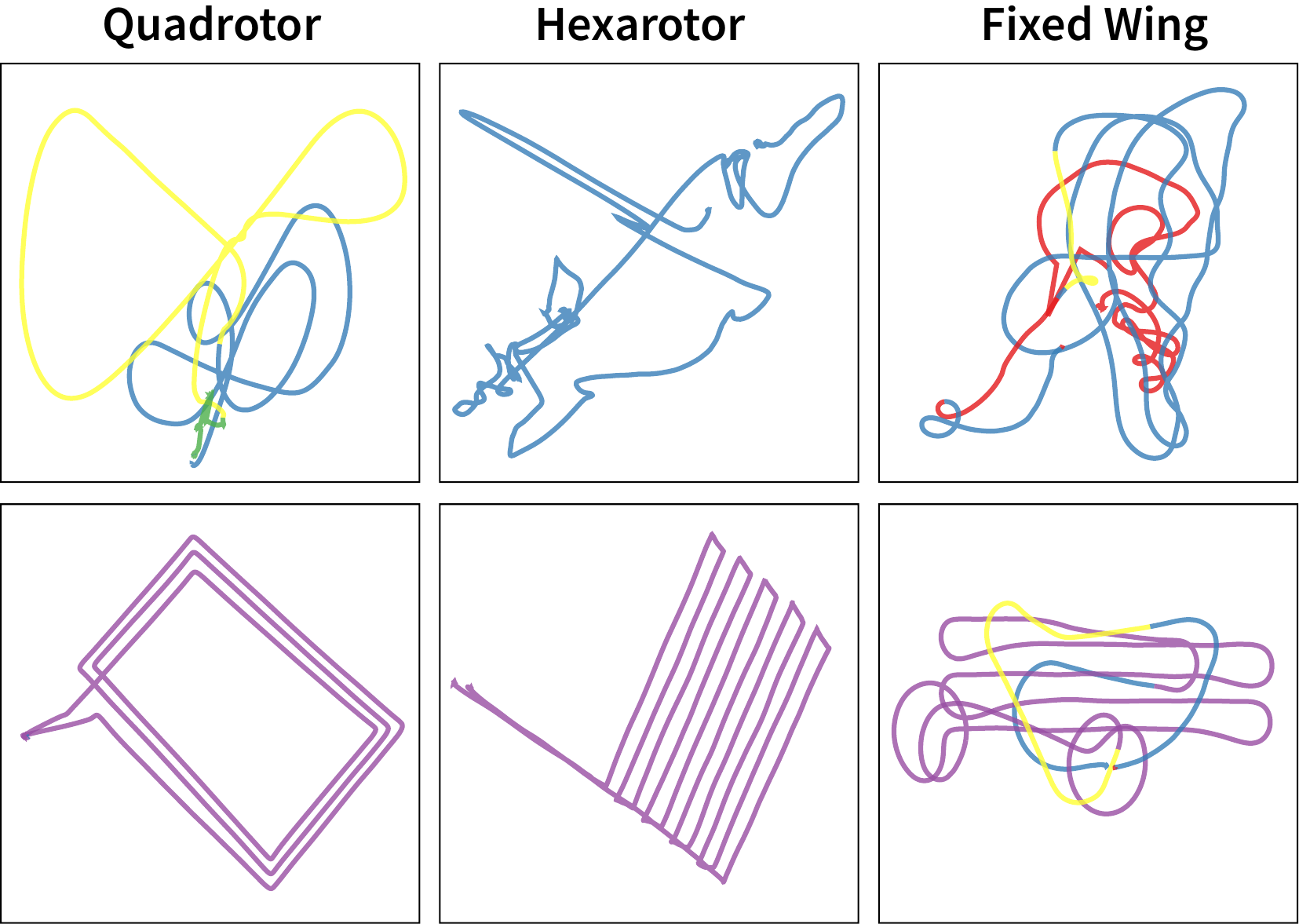}
\caption{Example of two flights for each vehicle type, where the top row represents manual and semi-guided flights and the bottom row represents mostly automated flights. 
The color represents different flight modes: \raisebox{0.1cm}{\fcolorbox{black}{Mission}{\rule{0pt}{0pt}\rule{0pt}{0pt}}} Auto, \raisebox{0.1cm}{\fcolorbox{black}{manual}{\rule{0pt}{0pt}\rule{0pt}{0pt}}} Manual, \raisebox{0.1cm}{\fcolorbox{black}{Stablized}{\rule{0pt}{0pt}\rule{0pt}{0pt}}} Stabilized, \raisebox{0.1cm}{\fcolorbox{black}{Position}{\rule{0pt}{0pt}\rule{0pt}{0pt}}} Position Control, and \raisebox{0.1cm}{\fcolorbox{black}{atitControl}{\rule{0pt}{0pt}\rule{0pt}{0pt}}} Altitude Control. Auto encompasses taking off or landing as well as any waypoint-based travel. The other color represents different manual control methods, from pure manual (red) to some assistance from the UAV (the other colors). Fixed-wing vehicles have the same characteristics when flying a mission or manually being flown. On the other hand, multirotor vehicles' flight characteristic change depending if their auto, straight direct lines, or manual, more curved flight paths.}
\label{fig:ExampleFlights}
\end{figure}

The rest of the paper is organized as follows.
\Cref{sec:related_work} reviews the literature related to UAV type prediction, including the adjacent fields of maritime anomaly detection and time series classification.
We then in \Cref{sec:data} detail the PX4 data we use and the differences between three key UAV types: quadrotors, hexrotors, and fixed-wing, see \cref{fig:ExampleFlights} for an illustration of these different drones.
\Cref{sec:methodology} covers our methodology for classifying these types, including machine learning model construction and data engineering.
We explain how to train the model in \Cref{sec:training_evaluation}.
Then, we describe our experiments for evaluating model performance (\Cref{sec:experiments}), analyze and discuss the results (\Cref{sec:results_discussion}), and finally address the limitations of our approach (\Cref{sec:limitations}). Lastly, we have a supplemental paper that includes additional tables and figures located in a public repository (\url{https://osf.io/mnsgk/}).

\section{Related Work}
\label{sec:related_work}

Using machine learning to classify UAVs based on flight data is a relatively nascent field, and existing work is mainly focused on anomaly detection.
Thus we also searched for relevant techniques from the related domain of maritime anomaly detection and time series classification more broadly.

\subsection{UAV Anomaly Detection}

Most of the limited research on machine learning with UAVs is focused on anomaly detection, and thus is limited to smaller datasets that include coded anomalies. Park et al.\ \cite{thresholding} used an autoencoder to determine if a flight is faulty based on whether a threshold is met when reconstructing the vector representations.
However, the datasets used are not sufficient for our task as we need a variety of flights flown by different types of UAVs to create a robust classifier.
The reconstruction loss between safe and faulty states would be higher, indicating an anomaly.
The two datasets used were the ALFA \cite{alfa} and the UAV attack \cite{UAV-Attack} datasets consisting of 47 flights and a single flight, respectively.
The ALFA dataset was from flights running ArduPilot whereas the UAV Attack dataset used PX4.
They generated additional samples by splitting the single flight into multiple chunks.
Chowdhury et al.'s \cite{thresholding2} approach to anomaly detection involves an autoencoder on IMU data using Rosbag files.
They used the files from UAVs that ran ArduPilot and collected compressed sensor data and images at respective time frames.
Their autoencoder portion is quite similar to Park et al.'s approach mentioned above.
Since our focus is more on flight logs and the data captured during each flight rather than images, much of their work is not applicable.

Lang et al.\ \cite{deeplearning} used an LSTM to determine whether a flight is faulty by checking if the prediction and the correct value's residuals are within an uncertainty range. 
Their LSTM approach better incorporates the temporal element of flight data, especially due to their use of a window length optimization technique that leverages the recent past to help with future predictions.
Our approach obviates the need for sliding windows by using the complete flights, 
Additionally, the structure of Lang et al.'s network changes throughout training because they only take one sample at a time; we preprocess our data to be trained in batches, maintaining a consistent structure in the LSTM.

An alternate approach to simplifying time series flight logs was proposed by Bronz et al.\ \cite{svm}, who concatenated multiple time steps as input to a support vector machine (SVM).
For example, when concatenating eight features and 20-time steps for each feature, a sequence of length 160 is generated. 
The data from their experiments are extracted from flights using the Paparazzi Autopilot System (PAS), but is not relevant is our case since there is not enough data and variety in the types of drones.
Rather than the typical one-class support vector machine (SVM) used for anomaly detection, Rahman et al.\ \cite{svm2} leverage a two-class SVM.
They utilized sensor measurements from PX4 flights---e.g.\ magnetometers, accelerometers, barometers, gyroscopes, and GPS---in addition to the duty cycle of each motor. 

Given the dearth of clean data with coded anomalies, many papers on UAV anomaly detection rely on simulations.
For example, Galvan et al.\ \cite{sim} simulated seven flight hours using PX4, Gazebo, and QGround-Control.
They then added simulated anomalies and detected them using neural networks.

Most of these papers focus on anomaly detection models and do not use data containing a significant number of flights flown by different types of UAVs.
These approaches also temporally aligned features to avoid needing timestamp sampling techniques.
In contrast, our work utilizes the full PX4 Flight Review dataset, which required more in-depth sampling and feature selection procedures to handle all the raw data.

\subsection{Maritime Anomaly Detection}

Machine learning using maritime data is highly relevant for our discussion, as both UAV and ship data incorporate a vehicle position that changes over time.
Research in the maritime domain is more mature and there is readily-available (and mostly correct) data from the Automatic Identification System (AIS) \cite{ais}.

A survey paper by Sidibé and Shu \cite{ais} provided us with an overview of anomaly detection with maritime vessels.
Lei \cite{maritimetraj} first simplifies the path of a vessel using grids and identifies different types of trajectories.
An anomaly can be detected by measuring the degree of suspicion for each trajectory and using thresholds.

Radon et al.\ \cite{context} partitioned given vessel tracks into different segments.
A clustering algorithm using comparison metrics such as dynamic time warping is applied to create clusters of similar segments.
An anomaly is detected if a segment does not fall into one of these groups based on a threshold.

Similarly, Laxhammar and Falkman \cite{lof} used the latitude and longitude position of vessels and applied local outlier factors to find anomalies in sub-trajectories or a portion of the complete path taken by a vessel. 
They also included a more general conformal anomaly detector that is capable of detecting false alarms fairly well and improved the computational efficiency of their previous approach \cite{lof2}. 
Jiang et al.\ \cite{auth} used behavior modeling and trajectory prediction using classical techniques involving Kalman filters to predict the illegal behavior of UAVs and leveraged the trusted authentication of the pilots to assist in their detection.
They focus more on having a UAV security platform for real-time detection and monitoring rather than creating a predictive model using UAV flights.

Most of the research on machine learning with maritime data deals with anomaly detection and trajectory prediction.
For this purpose, longitudinal and latitudinal positions are usually the only features used due to the limited movement capabilities of vessels.
The source and destination of the maritime vessel are also usually known ahead of time, which simplifies representation of movement and aids in detecting maritime vessel anomalies.
But it is not sufficient for detecting different types of UAVs based on the obvious differences in their mechanical capabilities.

\subsection{Time Series Classification}
To help us refine our goal of predicting UAV types, we explored existing literature on time series classification in general.
When dealing with time series data, there are both univariate and multivariate forms with different algorithms for each.
Univariate time series classification is much more common since classical machine learning models and neural networks can be used.
Yan and Oates \cite{tsdnn} used multilayer perceptrons (MLPs), fully convolutional neural networks (CNNs), and ResNet on 44-time series datasets from the UCR data repository, which includes domains such as medicine and biology. 
For multivariate time series classification, Seto et al.\ \cite{tsdtw} applied a modified dynamic time warping algorithm to classify human activities. 
Zheng et al.\ \cite{tscnn} used multi-channel deep CNNs for multivariate time series classification.
They acknowledged that the UCR data repository mostly contained univariate time series data and is comprised of primarily small datasets, which do not work well with CNNs.
Instead of looking for a different dataset, they combine two single datasets of left and right thorax heartbeats.
Since multivariate time series data can be treated like 2D images, CNNs are viable.
CNNs may be useful with sequential data when sequences are quite long.
They are often used to avoid exploding gradient issues during the backprogagation process of recurrent neural networks (RNNs).

The closest work that aims at UAV classification using temporal sequences is research that involves the detection of UAVs based on spectrograms extracted from radar, radio frequencies, or acoustic fingerprints.
Molchanov et al.\ \cite{microdoppler} used micro-doppler signatures from radars to classify 11 different types of flying objects, including fixed-wing UAVs, helicopters, and quadrotors.
Classical ML techniques such as support vector machines and naive bayes were employed to classify the different classes.
Mendis et al.\ \cite{radar} extracted Fourier transforms from microdoppler signatures and used deep belief networks to predict from three different types of flying objects (artificial birds, helicopters, and quadrotors).
However, there were only 70 instances of each class, not including 50 additional data-augmented instances.
Brooks et al.\ \cite{temporal} extracted 2D points from radar, fed them into wave equations, and returned a temporal series of points.
These series were predicted using CNN, RNN, and MLP models.
Although each of these approaches achieved greater than 85\% respective to their evaluation methods, only a small subset of flight characteristics was used for classification.
Location is primarily used and represented through radar or radio frequencies which do not fully capture the variations of flights.
Additionally, there has been research in identifying and detecting UAVs from the audio produced by their motion.
Al-Emadi et al.\ \cite{audio} applied CNNs, RNNs, and convolutional recurrent neural networks (CRNNs) to identify UAVs based on their acoustic fingerprints.
Similarly, Jeon et al.\ \cite{audio_2} used CNNS, RNNs, and Gaussian Mixture Models (GMMs) on the same task. 
They simulate an outdoor setting by artificially adding noise to UAV sound data then used deep learning to predict if that sample contained a UAV. The data used in these papers is more aligned with what would be relevant in an application setting. For this reason, we would want to move more in this direction. However, we do not have access to this type of data and used more readily available data that provides an ample amount of information about flights. 


Our proposed approach to detecting UAV type incorporates techniques from existing literature and various modifications to handle an unexplored dataset.
The main similarity between our approach and past time series classification research is that we preprocess our time series data in a way that can be classified using a many-to-one LSTM.
The architecture effectiveness has been demonstrated in a variety of previous research \cite{deeplearning, lstmfcn}.
Other than this similarity, we explore more logs, attributes, and drone types in the PX4 data and devise various novel techniques along the way to accommodate the unexplored data.

First, we use PX4 data much like many of the approaches in the UAV anomaly detection papers; however, we use significantly more data available in \cite{PX4}.
Second, our initial approach does not involve chunking the flights into multiple flights.
We also test more timestamp sampling techniques to represent a flight better rather than performing concatenations of time series data as done in \cite{svm} or randomly selecting single points from equal intervals as done in \cite{thresholding}.
Third, we explored more feature subsets and looked beyond the more typical features such as x, y, and z local positions and roll, pitch, and yaw angles.
Even though we do not exhaustively test all of the features available from PX4, we outline a procedure that can find potential feature candidates for future testing.
Lastly, we explore data augmentation and class sampling techniques as attempts to tackle the class imbalance problem.
Class imbalance is not an issue in the time series classification literature that we explored but is a huge problem in our setting since we intend to use as much PX4 data as possible.

\section{Data}
\label{sec:data}

For this work, we utilized PX4's flight data, one of the largest publicly available repositories of UAV data \cite{FlightReview}.
The data consists of flights uploaded voluntarily by users through a file format called ulogs.
The platform contains various UAVs and a wide variety of different firmware and software. 
Selecting a flight supplies an in-depth look at variables such as altitude, roll, and pitch angle, to name a few over time.
A 2D map is also provided, categorizing the flight into different modes of travel, such as manual, stabilized, or mission.

\subsection{Data cohesion}
\begin{figure}[t]
\includegraphics[width=.95\linewidth]{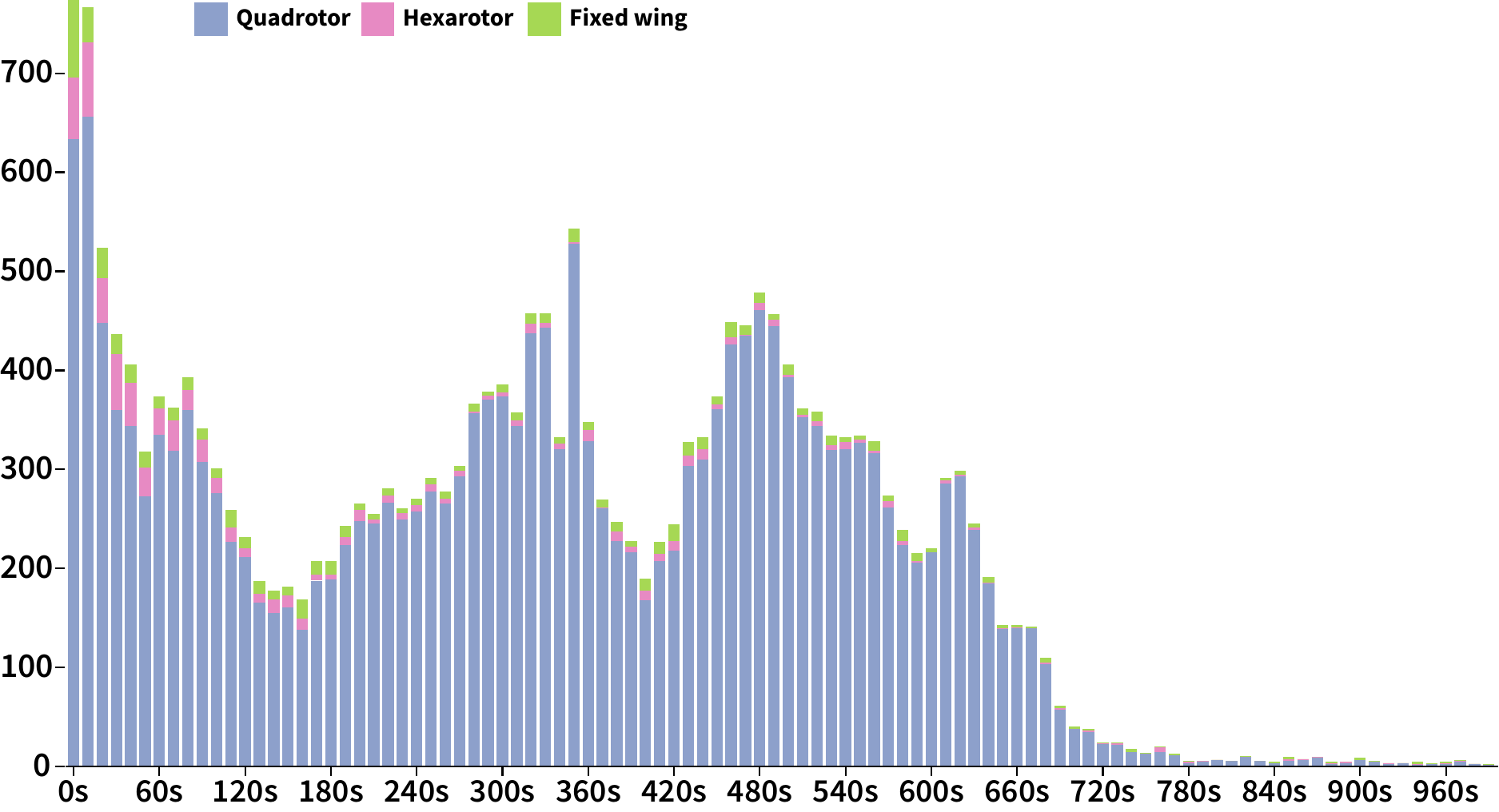}
\caption{A stacked bar graph showing the quantity of each UAV type for flights up to 1000 seconds. The x-axis is the duration of the flight binned in 10-second increments. Quadrotors make up most of the UAV types. Almost all flights are less than 10 minutes long with a majority under 30 seconds since PX4 review is often used to test UAVs.}
\label{fig:VehicleDuration}
\end{figure}

Having a comprehensive dataset with various types of flights and UAVs is of incredible value in this domain. 
However, the issue of data cohesion arises because of these variations. 
Individuals have their own goals and purpose for uploading their ulog to the website.
For instance, there is a significant spike of flights that are less than 30 seconds, shown in figure \cref{fig:VehicleDuration}, indicating these are probably testing logs to confirm everything is working properly with the vehicle.
These varied individual goals may generalize poorly to the goals of the flights that we want to detect. 
For example, a fixed-wing UAV flown at a low level for testing purposes would drastically differ from long-range, high-altitude flights in a surveillance task.
As a result, we need to create a robust model capable of learning the behavior of sensors in relation to the different usages of quadrotors, hexarotors, and fixed-wing UAVs.

\subsection{Differences Between Quadrotors, Hexarotors, and Fixed-Wing UAVs}

Understanding the mechanical differences between quadrotors, hexarotors, and fixed-wing UAVs is crucial for selecting the appropriate features and interpreting results.
Quadrotors and hexarotors are both multirotors, which can make sharp turns and move vertically. This mobility has a drawback of being very power inefficient relative to the other vehicle types and limits flight times to around 20--30 minutes.
The respective use cases will likely change as the number of rotors in a multirotor increases.
More motors allow for increased payload capacity and shifts the use case to transportation, whereas fewer rotors are better for inspection and aerial photography, as smaller UAVs can fit into tighter spaces.
The fixed-wing UAV is designed to function similarly to an airplane, having one rigid wing to provide lift.
Its rigid wing allows for a more power-efficient method of staying in the air, allowing it to cover longer distances and carry more weight.
Although fixed-wing UAVs can stay aloft for 16 or longer hours, pilots cannot place them in hover mode, making them more challenging to operate. The alternative is to fly them in a pattern such as circle, which can be inconvenient. 

Understanding the difference between the flight abilities of multirotor and fixed-wing UAVs gives us an intuition that a classifier could distinguish between two types of drones.
Furthermore, if we look at \cref{fig:ExampleFlights}, we can see the topological difference between the different UAV types.
Multirotor have different flight patterns depending on whether it is waypoint-based flying (Shown in purple) or manual (displayed in the rest of the colors). 
Multirotor in missioned-based flights make sharp turns going directly from one waypoint to the next.
When flown in manual mode, the paths turn more curved as that is the more natural form of flying.
On the other hand, fixed-wing UAVs, independent of the flight mode, have a slower turn, producing a more curved path, as shown in the flight's multiple loops.
These patterns might indicate that when flight mode is a mission type, the system will have an easier time telling multirotor and fixed-wing flight apart.
Telling a quadrotor and hexarotor apart seems more tricky, as the flight characteristics look visually similar. 

\subsection{Data Filtering}

For our classification task, we downloaded every log but only kept the quadrotor, hexarotors, and fixed-wing UAVs. Other types of drones in the PX4 data include octorotors and ground rotors, which were not used because we felt that having two similar (quadrotors and hexarotors) and different (quadrotors and fixed-wing UAVs) types of drones would be sufficient.
Figure \ref{fig:ExampleFlights} shows two flight examples for each vehicle's type.
We initially chose quadrotor and fixed-wing UAVs due to distinct flying characteristics as proof of concept that our method could detect the difference between these vehicle types.
We then expanded the classifier to include hexarotor to see if it was possible to predict vehicle types that were considerably more similar (hexarotor and quad rotors).
This extraction resulted in 29,362 flights after removing the logs that had parsing errors from the library we used. 
Of these flights, 26,706 logs were from quadrotors, 1,332 from hexarotor, and the remaining 1,324 were from fixed-wing UAVs. 
Clearly, there is a significant imbalance in the number of quadrotor flights compared to the number of hexarotor and fixed-wing flights. This imbalance can be explained by the fact that quadrotors are cheaper and have recently become popular for recreational use. The amount of data is further reduced in each class after feature selection, see \cref{sec:feat_select}.
This imbalance may lead to a challenging classification task if the appropriate methods are not used.

\section{Methodology}
\label{sec:methodology}

We first explore the potential models to use for time series classification before handling the data.
To narrow down the options, we looked into the models used in literature and rationalized why certain models are less ideal for our approach.
Doing this step first allows us to determine how the data needs to be engineered to fit the input of the model.
The unstructured nature of the data requires some form of feature selection and filtering before we can engineer the data.
For this reason, we move on to feature selection after the model.
The final step is to engineer the data to fit our model, which involves exploring different timestamp sampling techniques.

\subsection{Model}
We can better understand how to engineer our data to be used for classification by beginning with the model. 
We considered the fact that even though this data is time series, we are not performing a forecasting task.
As a result, we are more interested in models that can learn the temporal nature of the data and output a single value that represents the predicted class.
One possibility is using dynamic time warping (DTW), which combines Euclidean distance and dynamic programming to compare time series data, and is typically used for univariate data \cite{tsdtw}. 
Even though there are modifications to this algorithm to accommodate multivariate data, this method is time inefficient with the size of our data since it compares time series data in quadratic time.
Exploring DTW is not ideal since our data is not well explored and requires a significant number of training experiments. 

We could use more classical ML models by manipulating the data such as SVMs, Naive Bayes, or decision trees.
However, the general reason these models are not used is the high number of features. 
Using these models would require concatenating each feature, which neglects the time component and may also suffer from a high number of dimensions \cite{dim}.
Additionally, SVMs do not support multi-class classification and are designed for binary classification.
The workaround is to create multiple binary classifiers, which for our case, makes it difficult to compare the effects of a majority class's distribution to other minority classes. 
It is possible to reduce these problems mentioned above by using deep learning models.

An example of a potential deep learning method would be multi-channel CNNs as performed in \cite{tscnn}.
Multi-channel CNNs involve breaking down a multivariate task into several univariate tasks.
Then, a series of feature extractors and hidden layers are used with a final MLP for classification. 
However, our primary objective is to see how changes to data preprocessing and tackling class imbalance affect performance instead of focusing heavily on the model's architecture.
We want our pipeline for classifying UAV types to serve as a starting point in which models can be swapped out and tested.

RNNs are used frequently in time series tasks, especially for forecasting.
Since time series data is inherently sequential and order matters, the memory capability of RNNs is crucial.
Therefore, we consider RNNs as an intuitive model for our data.
Our pipeline uses Long-Short Term Memory networks (LSTMs), a variant of RNNs.
LSTMs address the problem with vanishing gradient in RNNs that occurs much more frequently because of the number of time steps \cite{lstm}.
Solving this problem also leads to computational efficiency.
LSTMs have the downside of not being able to capture long-range dependencies.
However, our task does not require it for two reasons.
Specifically, a simple LSTM with a single layer and 128 LSTM cells was used for our experiments to great effect, see figure 1 in our supplemental material (\url{https://osf.io/mnsgk/}) for model illustration and \Cref{sec:training_evaluation} for results. 
This single layer is paired with linear classifier output.
By configuring this model to take in an input equal to the number of binned intervals, a many-to-one LSTM and classifier were created to classify if a flight sequence is from a quadrotor, hexarotor, or fixed-wing UAV.
Once we have decided on a model, we can focus on preparing the data for the LSTM, starting with feature selection.



\subsection{Feature Selection}
\label{sec:feat_select}
There are major challenges when handling the entire available set of raw PX4 flight logs. 
The data for each flight is comprised of topics, which are a collection of features derived from the same sensor \cite{PX4}.
For example, the topic `vehicle local position` will likely contain `x`, `y`, and `z` features relative to the starting position.
The PX4 data set contains over 325 of these topics. 
However, only 27 topics are contained in at least 60\% of the flight logs.

Another challenge is that we were unable to find consistency in these topics with a particular software version or hardware type.
In some instances, a topic missing makes sense, such as some fixed-wing lacking battery information, as a fixed-wing can run off of gas power.
On the other hand, some topics may vary in naming convention despite capturing the same data.
We learned that these topics have evolved over time; therefore, finding a common set of topics to aggregate a processed data set for model training and evaluation is difficult.
This characteristic of inconsistency also holds true for the features of each topic.

One possible technique for feature selection involves cycling through all permutations, such as sequential forward selection (SFS).
SFS finds a potential feature subset by gradually adding a single feature \cite{sfs}.
At each iteration, the best-performing subset is chosen, and we continue the process with the permutations for the remaining features. 
This approach, along with other feature selection methods, assumes that all features are equally contained across samples.
However, with our data set, that is not the case.
This data inconsistency creates instances where a combination of features can really limit the data's availability.
Therefore, we have to be careful in our feature selection.


It is unrealistic to exhaustively test all combinations, but at the same time, we do not want to constrain ourselves to a very small subset of features; therefore, we tried several possible feature subsets based on varying criteria.
This approach ensures that we explore more than just the features proposed in research related to our task and does not require cycling through various features that may not have any value in prediction. 

The first step of this process involves only checking topics if they are contained in at least 60\% of the logs.
Since many topics may be unique to specific logs, we can prune these less-common topics out. 
Additionally, even if logs have the same topics, they may not necessarily contain the same features.
Therefore, we applied the same technique of using only features contained in at least 60\% of logs.
These steps reduce the feature search space and allow us to maintain a reasonable dataset size for testing.

We begin with a base set of features from \cite{thresholding} to reduce the variability that comes from completely random features.
These features are the x, y, and z local positions as well as the roll, pitch, and yaw angles.
Next, we select the number of feature subsets we want, randomly select $n$ features without replacement from the pruned features, and manually remove some of the features.
Manual removal requires some assumptions to be made to simplify the process.
We consider features not related to motion and/or time as irrelevant.
For example, within the 'vehicle gps position' topic, there are features called 'satellites used' and 'jamming indicator'.
Both are more status-based features that can be manually removed.

Given these feature subsets, we can perform experiments with our model to give us a rough estimate of which features work best.
Details regarding the number of feature subsets tested and the features used are clarified in the experiments section (\cref{sec:experiments}).

We acknowledge that this approach is not exhaustive, and  that the quantity of data will change with each subset.
As a result, we recognize that the variations in performance using these feature subsets may be a result of the different derived class distributions.

\subsection{Timestamp Sampling}

Before we can determine an approach to handle the variable durations, we first need to understand that this process is further complicated by the fact that there are multiple features per sample that will likely sample at different times and at different rates.
The most straightforward way of addressing this issue it to align these features temporally and apply zero padding to the beginning and end of shorter sequences. 
One downside to this approach is that the data may become very sparse since shorter sequences may become mostly zeros if much longer sequences exist.
This problem is not as severe in our task considering that we have a much smaller subset of features that are similar in beginning and end timestamps.
For example, the angular acceleration and local position sensor values are likely to be aligned because of their similarity.
Our approaches are based on zero-padding with modifications to determine which values are sampled from the aligned features. Once we assign a way to align the features, we can proceed with sampling timestamps to bin and summarize the data.

Timestamp binning refers to the sampling method used to convert the time series instances from variable to fixed length.
The simple approach is take the maximum and minimum of timestamp of each log and divide the timestamps into equally sized intervals called equal-width binning.
We produce a dataset of equal length time series data because the number of intervals are fixed. This local maximum and minimum timestamp per log ensures that each feature sampling lines up at the interval timestamps since there are multiple features per sample,. 
Having a higher number of intervals increases the resolution of each part of the flight.
However, the differences in durations make it difficult to find an appropriate number of intervals to accommodate these variations, which we will further discuss in our experiments.
We reduce the number of outliers that could skew the range of timestamps by performing this independently for each log \cite{disc}. 
Once we have these intervals, we need to obtain a single value to represent the values within each interval. 

There are drawbacks and benefits to the different approaches to obtaining a single value.
Park et al.\ assumed that a randomly selected value could represent a feature during a time window \cite{thresholding}.
In their setting, this approach makes more sense because of the short duration of each of their instances (30 seconds).
However, because of the longer durations of our flights and the variations, taking a random sample would not accurately represent a full interval.
Taking the maximum or the minimum within each interval suffers with sensitivity to outliers and may not adequately represents parts of a flight log with a significant amount of movement and change in directions.
We intend to sample data in a way that captures more information about the flight or aligns data between different flights.

Here we detail the two sampling approaches we compared to determine how to engineer the data to fit our model.
The first approach we refer to as average sampling, which divides each flight into an equal number of bins based on a local minimum and maximum timestamp for each flight.
The second approach we call fixed window average sampling, which essentially shrinks the number of timestamps used in the first approach to a specified time range.

\subsubsection{Average Sampling}
We first test average sampling to see if we can represent and summarize a flight better when compared to selecting a random single point.
Figure \ref{fig:equalAvg} shows a visual of how average sampling works.
Each flight has a local minimum and maximum as determined by the first and last timestamps of every features.
Since features may have different ranges, there will likely be empty space for shorter ranges.
However, we take the minimum and maximum and can divide each sample into an even number of intervals.
With the complexities involved in flight data, there is high potential in abstracting away details in a particular flight.
For these reasons, we see an opportunity in exploring more timestamp sampling methods based on our domain knowledge of flight patterns.



\subsubsection{Fixed Window Average Sampling}

\begin{figure}
    \centering
    \begin{subfigure}[t]{\linewidth}
        \centering
        \includegraphics[width=.95\linewidth]{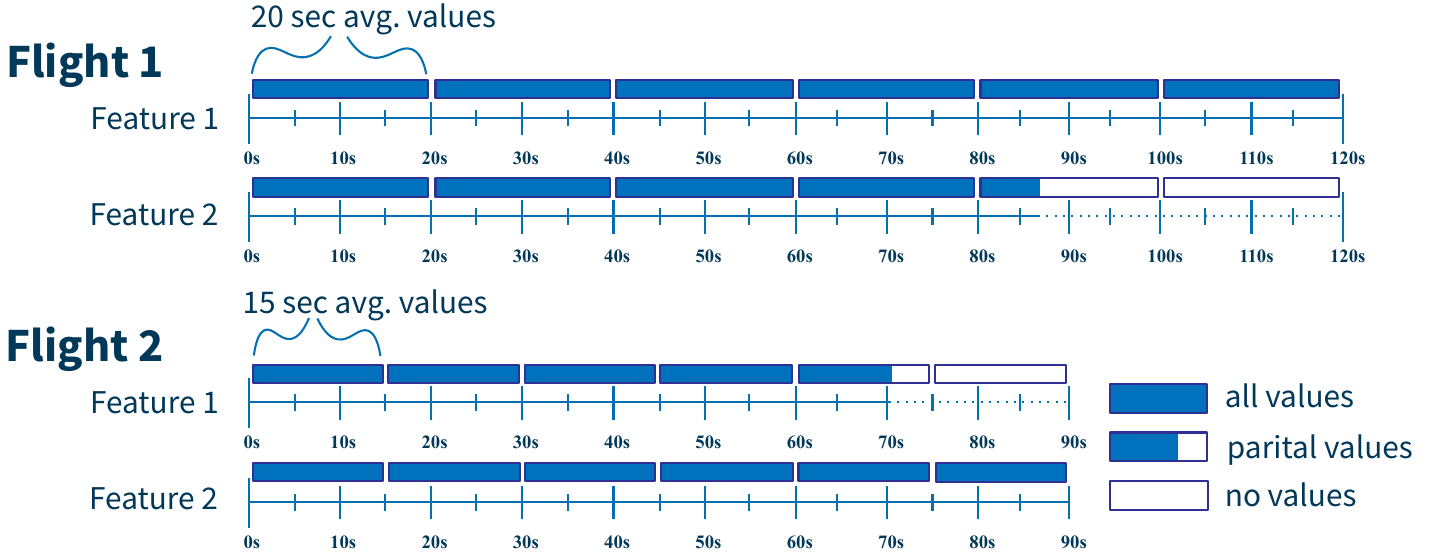}
        \caption{Average Sampling}
        \label{fig:equalAvg}
    \end{subfigure}
    \begin{subfigure}[t]{\linewidth}
    \centering
        \includegraphics[width=.95\linewidth]{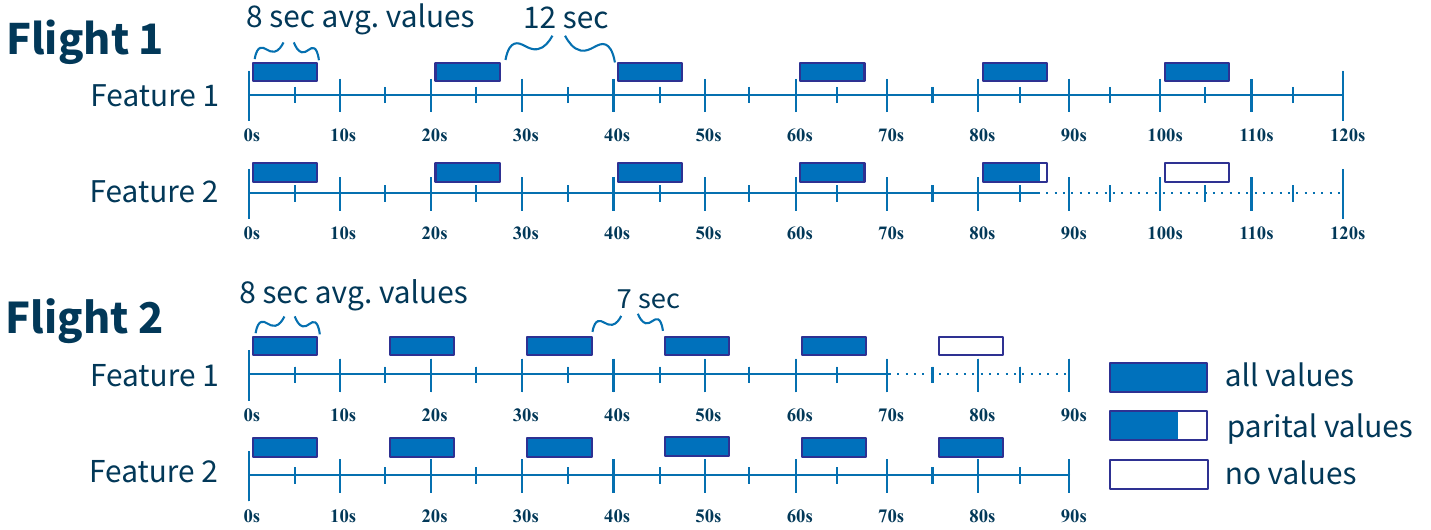}
        \caption{Fixed Window Average Sampling}
        \label{fig:windowAvg}
    \end{subfigure}
    \caption{The two figures illustrate how fixed window average and average sampling work on flights with different flight times. For average sampling, the amount of data averaged changes across flights, with shorter flights averaging a smaller portion of the data. Fixed window average averages the same amount of data across flights but varies the gaps. For each flight, the algorithm uses the longest feature for sampling. For any shorter feature, we sample partially or not at all to guarantee that the sampling is aligned within a flight.}
    \label{fig:timeSampling}
\end{figure}

Windowed averaging reduces and normalizes the amount of information compared across different flights. For example, as shown in Figure \ref{fig:windowAvg}, we can capture an 8 second window, which abstracts less data than averaging over an entire interval and acquires more data than sampling a single point. Averaging over the same duration means that the same durations are compared across different flights unlike the averaging approach which may average over more or fewer points depending on the length of the flights. Figure \ref{fig:windowAvg} shows that since we are using the same window size for each flight, the amount of space between the end of each window and the beginning of the next window may vary (i.e.\ 12 seconds vs.\ 7 seconds). This method introduces a new parameter to tune, the window duration, which will be discussed further in the experiments section. Selecting this value involves analyzing the overall duration of the flights for each class.

\section{Training and Evaluation}
\label{sec:training_evaluation}

We performed a standard neural network training procedure depending on a set number of epochs and batch size.
The standard classification loss, cross-entropy, was used alongside the Adam optimizer during gradient descent.
To obtain consistent results, we ran experiments using different training configurations until the results stabilized and provided consistent results.
Because of the heavy imbalance in the dataset and relatively quick training times, we also employed k-fold cross-validation.
We average the results across all folds and compute standard deviation to determine the consistency of the results.
Each fold produces a precision, recall, and f-score for the number of quadrotors, hexarotors, and fixed-wing UAVs identified correctly.
Although not necessary, averaging results across multiple folds of data ensures a better representation of performance and robustness in the few instances of hexarotor and fixed-wing flights.


We need to determine the ideal and appropriate results for the task.
First, we need to clarify the difference between precision, recall, and f-score.
Precision indicates the number of actual positive predictions out of the total number of predicted positives.
We would want precision to be high if we desired to reduce the number of false positives.
In the case of classifying multi-rotors vs.\ fixed-wing UAVs, we can look at the problem from the perspective of identifying potential threats.
Fixed-wing UAVs can typically travel farther distances and carry larger payloads when compared to the typical multi-rotors.
One implication of this characteristic is that the fixed-wing UAV has more opportunities due to range and may contain a larger harmful payload.
In this case, a false positive can be seen as an inconvenience that can lead to alarm fatigue.
However, recall indicates the number of positive predictions out of the total number of actual positives.
We would want recall to be higher if we wanted to reduce the number of false negatives or missing out on potential threats.

A similar argument can be made for quadrotors vs. hexarotors.
The additional rotors allow for transporting larger payloads.
These scenarios are based on our domain knowledge of the capabilities of these types of UAVs and may not completely align with other applications.
If we do not want to limit ourselves to a particular scenario, we can use f-scores.
F-scores indicate the harmonic mean between the precision and recall to indicate overall performance. These scores are averaged across the three classes in an unweighted manner known as macro averaging,
ensuring that the majority class does not completely dominate the minority classes \cite{macrof1}. We use this single score to summarize performance.

\section{Experiments}
\label{sec:experiments}

Our goal for our experiments is to determine if the processed flight data can be used to distinguish different types of UAVs.
We take it further by comparing different sampling methods and data augmentation techniques.
Exploring these methods not only helps improve results but can also translate over to different classification tasks involving this type of data.

\newcommand\items{3}   
\begin{table}[!htp]\centering
\begin{threeparttable}[!htb]\centering

\noindent\begin{tabular}{cc*{\items}{|c}|}
\multicolumn{1}{c}{} &\multicolumn{1}{c}{} &\multicolumn{\items}{c}{Predicted Class} \\
\multicolumn{1}{c}{} & 
\multicolumn{1}{c}{} & 
\multicolumn{1}{c}{Quadrotor} & 
\multicolumn{1}{c}{Fixed-Wing} & 
\multicolumn{1}{c}{Hexarotor} \\ \cmidrule{3-5}
\multirow{\items}{*}{\rotatebox{90}{True Class}} 
&Quadrotor  & 12742   & 56  & 83   \\ \cmidrule{3-5}
&Fixed-Wing  & 124   & 278  & 10   \\ \cmidrule{3-5}
&Hexarotor  & 214   & 15   & 118   \\ \cmidrule{3-5}
\end{tabular}


\end{threeparttable}

\caption{Combined confusion matrix for every fold for trial 1, see \cref{tab:TimestampSampling} for trial reference. The diagonal values indicate the instances that were predicted correctly and the other values were the misclassifications. We can see that a majority of hexarotors were classified as quadrotors and a good portion of fixed-wings were also classified as quadrotors.}
\label{tab:confMat}
\end{table}

\subsection{Preliminary Experiments}
We perform some experiments prior to comparing timestamp sampling and class imbalance techniques.
These experiments help establish a starting point for our compared techniques. 

\subsubsection{Preliminary Experiment 1---Feature Selection}
As discussed in our methodology selection, we employ an improvised feature selection approach to explore more of the available data from PX4's flight review.
To reiterate, a feature subset from \cite{thresholding} gave us the base set of features in addition to a few more features.
The base set of features with these additional features gives us the baseline features.
For our experiments, we looked into three additional feature subsets.
We chose the feature subset that resulted in the best performance and proceeded with our other experiments. To view the features tested and their resulting class distributions, reference table I and table II shown in the supplemental section (\url{https://osf.io/mnsgk/}). 

\subsubsection{Preliminary Experiment 2---Achieving Consistent Results}

We do not compare our model to existing state-of-the-art models; therefore, we first attempt to get consistent results obtained when the classification pipeline is completed.
From preliminary experiment 1, we found that the features used in \cite{thresholding} led to the best performance; however, we noticed that performance varied drastically depending on the training and model hyperparameters across different folds.
Therefore, we tuned these hyperparameters until results stabilized across 10 folds of data. 

\subsection{Main Experiments}

After these preliminary experiments, we define the components of the pipeline that are fixed. The preliminary experiments gave us the following features: the x, y, and z local positions, the roll, pitch, and yaw body angles, the throttle, altitude, and battery temperature. They also gave us the training and model hyperparameters such as a 0.001 learning rate, Adam optimizer, and a single hidden layer LSTM with 128 nodes. We can then proceed with these features and hyperparameters.

The first experiment involves trying different timestamp sampling techniques and analyzing their effects on model performance.
We begin with timestamp sampling since choosing a technique does not require any changes to the distribution of classes.
Fixing variables and choosing this experiment order still effectively shows the benefits and drawbacks of each tested technique.
Furthermore, this approach reduces the variability that comes with changing many variables at once.
We test a set of hyperparameters for each technique in the following experiments.

\subsubsection{Experiment 1---Timestamp Sampling Variations}
Each of the timestamp sampling variations mentioned in the methodology section is tested alongside various hyperparameters.
To summarize, there are two techniques in total.
Equal-width average sampling has a single hyperparameter that determines how many intervals the flight is divided into.
Equal-width windowed averaging has two hyperparameters, the number of intervals and window duration at each interval. We performed some analysis on the duration of flights to rationalize our testing space.

Further analysis of the duration of flights helps us narrow down the parameters to test for the different techniques.
Over 99\% of the logs were under 60 minutes.
As expected, the average durations of multirotors were lower than those of fixed-wing UAVs.
Multirotors had an average flight duration of 5.56 minutes, whereas fixed-wing UAVs remained in flight for almost 2 minutes longer at 7.48 minutes.
We can assume that fixed-wing UAV flights were not much longer because most logs in the database were the result of simple flights; therefore, we expect most flights to have around the same average flight times.

By using this information, we can justify some our hyperparameter configurations.
For example, if we have 50 intervals, an 8-minute flight would be divided into roughly 10 seconds chunks. 
In this case, the window at each interval would have to be less than 10 seconds. 

We acknowledge that some flights may be much shorter than the average, meaning the windows may end up filling the whole interval. However, for the sake of consistency in our experiments, we proceeded with the following parameters based on our justification from the previous paragraph. 50, 200, and 500 intervals are used for each technique and 2, 5, and 10-second windows are used for the windowed averaging technique, see \cref{tab:TimestampSampling} and \cref{tab:classInbalance} for each trial configuration.


\begin{table*}[!htp]\centering
\begin{threeparttable}[!htb]\centering
\scriptsize
\begin{tabular}{lrrrrrrrrrrrrr}\toprule
& & &\multicolumn{3}{c}{Quadrotor} &\multicolumn{3}{c}{Fixed-Wing} &\multicolumn{3}{c}{Hexarotor} & \\\midrule
&Trial Ref.\ Number &Parameter &Precision &Recall &F-Score &Precision &Recall &F-Score &Precision &Recall &F-Score &Macro F-Score \\
\arrayrulecolor{gray}\hline
&& Num.\ of Intervals & & & & & & & & & & \\ 
\hline
Average Sampling &1 &50 &\ul{\textbf{97.42}} &\ul{\textbf{98.92}} &\ul{\textbf{98.16}} &\ul{\textbf{80.51}} &\ul{\textbf{67.46}} &\ul{\textbf{73.15}} &\ul{\textbf{57.08}} &\ul{\textbf{34.03}} &\ul{\textbf{42.15}} &\ul{\textbf{71.15}} \\
&2 &200 &97.2 &99.36 &98.26 &85.79 &69.17 &76.34 &62.35 &24.5 &34.61 &69.74 \\
&3 &500 &96.9 &99.46 &98.17 &85.59 &12.96 &21.32 &63.46 &12.96 &21,32 &65.83 \\
\hline
& &Num.\ of Intervals, Win.\ Size (sec)  & & & & & & & & & &  \\
\hline
Fixed Window &4 &50, 2 &96.05 &99.02 &97.51 &67.52 &51.43 &57.91 &52.33 &5.49 &9.69 &55.03 \\
Average Sampling &5 &50, 5 &96.16 &99.22 &97.66 &73.37 &54.33 &62.15 &47.33 &4.61 &8.13 &55.98 \\
&6 &50, 10 &\textbf{96.18} &\textbf{99.33} &\textbf{97.73} &\textbf{78.48} &\textbf{57.25} &\textbf{65.82} &\textbf{50} &\textbf{4.02} &\textbf{7.18} &\textbf{56.91} \\
&7 &200, 2 &95.87 &99.2 &97.51 &68.77 &45.56 &54.36 &46.71 &4.62 &8.21 &53.36 \\
&8 &200, 5 &95.96 &99.32 &97.61 &72.9 &47.76 &57.57 &48.52 &5.49 &9.71 &54.96 \\
&9 &200, 10 &96 &99.21 &97.58 &71.88 &52.38 &60.2 &23.5 &2.58 &4.63 &54.13 \\
&10 &500, 2 &95.77 &99.25 &97.48 &68.65 &42.71 &52.41 &45.24 &3.45 &6.15 &52.01 \\
&11 &500, 5 &95.89 &99.36 &97.59 &72.17 &48.76 &57.85 &43.44 &2.01 &3.81 &53.09 \\
&12 &500, 10 &95.84 &99.38 &97.58 &72.27 &4706 &56.84 &25 &1.14 &2.14 &52.18 \\
\bottomrule
\end{tabular}

\smallskip
\caption{The results after applying different timestamp sampling techniques. \textbf{Bolds} represent the highest macro f-score for each approach and \ul{underlines} represent the highest macro f-score for each experiment. Average sampling gives the best overall performance as shown by the macro f-score. Even though the results between using 50 and 200 intervals for average sampling is not much different, we see that performance begins to decline as we increase the number of intervals.}
\label{tab:TimestampSampling}
\end{threeparttable}

\end{table*}

\subsubsection{Experiment 2---Fixing Imbalanced Classes}

As mentioned previously, there is a large imbalance in the number of quadrotors compared to the number of fixed-wing flights and hexarotors.
We address this problem after determining the ideal timestamp sampling technique.
This technique and the appropriate hyperparameters are fixed to proceed with the class imbalance experiment.
Commonly used techniques for class imbalance are data augmentation, oversampling, and undersampling. 

The Tsaug library lets us apply data augmentation with tunable parameters to time series data \cite{tsaug}.
These parameters include cropping random subsequences, drifting the signal, and reversing sequences.
By adding these parameters, artificial data can be generated based on existing data.
We apply this library to the fixed-wing and hexarotor classes to create a better balance in the class distribution.

The Imblearn library offers a range of oversampling and undersampling techniques \cite{imblearn}.
One way to perform these techniques is to randomly oversample or undersample from either the majority or minority class with an assigned ratio.
Alternatively, more ML-based techniques such as cluster-centroid undersampling or synthetic minority oversampling technique (SMOTE) can be used.
Cluster centroid undersampling replaces instances from the majority class with the centroid of the K-means algorithm \cite{imblearn}.
SMOTE oversampling generates synthetic samples of the minority class by essentially interpolating samples from a data point and its nearest neighbors \cite{oversample}.

For our experiments, we either reduce the number of instances in the majority class or increase the number of instances in the minority classes at the same percentage.
For example, when undersampling, we decrease the number of quadrotors by 25\%, 50\%, and 75\%, either with random undersampling or cluster centroid undersampling.
Similarly, when we oversample either with random oversampling or SMOTE, we increase both the number of fixed-wing and hexarotor instances by 150\%, 200\%, or 250\%.
Table II in the supplemental section (\url{https://osf.io/mnsgk/}) shows how the class distributions change with these increases and decreases.
To ensure that we are not introducing additional bias, we ensure that the test folds are not modified with any form of augmentation or interpolated sampling.

\begin{table*}[!htp]\centering

\begin{threeparttable}[!htb]\centering
\scriptsize
\begin{tabular}{lrrrrrrrrrrrrr}\toprule
& & &\multicolumn{3}{c}{Quadrotor} &\multicolumn{3}{c}{Fixed-Wing} &\multicolumn{3}{c}{Hexarotor} & \\\midrule
&Trial Reference Number &Parameter &Precision &Recall &F-Score &Precision &Recall &F-Score &Precision &Recall &F-Score &Macro F-Score \\ 
  \arrayrulecolor{gray}\hline
& &Increase \% & & & & & & & & & & \\  
\hline
Data &13 &150 &98.23 &94.22 &96.17 &81.06 &59.65 &68.36 &26.05 &70.94 &37.53 &67.36 \\
Augmentation &14 &200 &\ul{\textbf{98.09}} &\ul{\textbf{94.41}} &\ul{\textbf{96.28}} &\ul{\textbf{84.85}} &\ul{\textbf{66.67}} &\ul{\textbf{74.67}} &\ul{\textbf{25.81}} &\ul{\textbf{70.59}} &\ul{\textbf{37.8}} &\ul{\textbf{69.58}} \\
&15 &250 &98.08 &94.29 &96.14 &79.57 &57.72 &66.69 &25.09 &68.87 &36.73 &66.52 \\ 
\hline
& &Increase \% & & & & & & & & & & \\ 
\hline
Random &16 &150 &\textbf{98.21} &\textbf{94.91} &\textbf{96.52} &\textbf{79.82} &\textbf{69.62} &\textbf{73.97} &\textbf{28.27} &\textbf{61.98} &\textbf{37.18} &\textbf{69.22} \\
Oversampling &17 &200 &98.61 &93.3 &95.88 &78.61 &69.13 &73.24 &24.48 &75.49 &36.84 &68.65 \\
&18 &250 &98.7 &92.26 &95.37 &77.66 &65.75 &70.86 &21.88 &78.12 &34.1 &66.78 \\
\hline
& &Decrease \% & & & & & & & & & & \\ 
\hline
Random &19 &25 &\textbf{98.91} &\textbf{91.79} &\textbf{95.21} &\textbf{71.74} &\textbf{72.59} &\textbf{71.53} &\textbf{22.13} &\textbf{78.71} &\textbf{34.39} &\textbf{67.04} \\
Undersampling &20 &50 &98.68 &92.86 &95.68 &62.93 &71.38 &66.51 &24.65 &72.34 &36.53 &66.24 \\
&21 &75 &99.06 &91.17 &94.95 &67.07 &75.74 &70.56 &21.85 &81.59 &34.39 &66.63 \\
\hline
& &Increase \% & & & & & & & & & & \\ 
\hline
SMOTE &22 &150 &\textbf{98.05} &\textbf{95.62} &\textbf{96.81} &\textbf{77.04} &\textbf{69.42} &\textbf{72.67} &\textbf{30.95} &\textbf{53.34} &\textbf{36.39} &\textbf{68.62} \\
Oversampling &23 &200 &98.74 &92.8 &95.68 &71.08 &73.77 &72.22 &22.9 &72.36 &34.73 &67.54 \\
&24 &250 &98.81 &92.75 &95.68 &72.37 &71.58 &71.83 &23.88 &77.55 &36.38 &67.96 \\
\hline
& &Decrease \% & & & & & & & & & & \\
\hline
Cluster Centroid &25 &25 &\textbf{98.49} &\textbf{93.84} &\textbf{96.11} &\textbf{69.94} &\textbf{67.47} &\textbf{68.35} &\textbf{24.71} &\textbf{68.65} &\textbf{36.26} &\textbf{66.9} \\
Undersampling &26 &50 &98.66 &92.39 &95.41 &64.13 &68.93 &65.84 &23.21 &73.27 &35.03 &65.42 \\
&27 &75 &98.98 &90.42 &94.49 &57.13 &67.71 &61.69 &21.11 &82.76 &33.51 &63.23 \\
\bottomrule
\end{tabular}

\smallskip
\end{threeparttable}
\caption{The results after addressing the class imbalances using each of the data augmentation and class sampling techniques. \textbf{Bolds} represent the highest macro f-score for each approach and \ul{underlines} represent the highest macro f-score for each experiment. Performance remains consistent across the techniques and configurations, but there are tradeoffs with precision and recall as addressed in table 4. For the oversampling and undersampling, the performance begins to drop as the changes in distribution are more drastic as shown by the fact that the top performer for each technique is the first configuration.}
\label{tab:classInbalance}
\end{table*}

\begin{table}[!htp]\centering
\begin{threeparttable}[!htb]\centering
\scriptsize
\begin{tabular}{lrrrrrr}\toprule
& &\multicolumn{2}{c}{Fixed-Wing} &\multicolumn{2}{c}{Hexarotor} \\\midrule
&Trial &Precision &Recall &Precision &Recall \\
\hline
Without Changes to Distribution &1 &80.51 &67.46 &57.08 &34.03 \\
With Changes to Distribution &21 &67.07 &75.74 & & \\
With Changes to Distribution &27 & & &21.11 &82.76 \\
\hline
Precision and Recall Tradeoff & &-13.44 &+8.28 &-35.97 &+48.73 \\
\bottomrule
\end{tabular}

\smallskip

\end{threeparttable}
\caption{The tradeoff in precision and recall when changing the distribution of classes using over and undersampling. Depending on the class sampling technique used (refer to the trial numbers), there is chance of increasing precision at the cost of recall and vice-versa with increasing recall.}
\label{tab:percisionRecallTradeoff}
\end{table}


\section{Results and Discussion}
\label{sec:results_discussion}

In the following section, we present and discuss the results of our experiment.
\subsection{Results}

The results of our two main experiments are shown in Tables \ref{tab:TimestampSampling} and \ref{tab:classInbalance}. For Experiment 1, average sampling using 50 intervals led to the highest performance. After averaging across 10 folds, we reach f-scores of 98.16\%, 73.15\%, and 42.15\% for the quadrotor, fixed-wing, and hexarotor classes, respectively. The overall performance of this technique can be summarized with the macro f-score, which was 71.15\%. 

We proceeded with this sampling method and number of intervals for Experiment 2, in which performance did not improve across all of our attempts to address class imbalance. If we look at the results of Experiment 2 independently of the first experiment, we achieved the best performance using data augmentation with a 200\% increase to the minority class. This configuration led to f-scores of 96.09\%, 74.67\%, and 37.8\%, for the quadrotor, fixed-wing, and hexarotor classes, respectively. 

There is a challenge of qualifying these results since there is no baseline of comparison, but we can think from the perspective of choosing only the quadrotor class since it is the majority or randomly guessing the UAV type. This approach would only give a macro f-score of 32\% since all the minority classes are not predicted. By guessing each class with a 1/3 probability, we on average reach a macro f-score of 20\%. Guessing each class based on the distribution of classes would yield similar results to guessing quadrotor each time.


\subsection{Analysis and Discussion}

The analysis performed on these results is done both quantitatively and qualitatively.
While we can simply look at the resulting performance of each modification, there are other factors to consider, such as speed and ease of implementation.
These factors are important when using such models in real-life settings, especially if performance is relatively consistent across techniques.
We can make more accurate comparisons of different techniques by analyzing the averages across ten folds of data.

\subsubsection{Feature Selection}
After testing different subsets of features, we found that the features provided in \cite{thresholding}, which we will refer to as the baseline features, led to the best performance. 
However, we do have to factor in the variations in class distributions.
Since the baseline features resulted in much fewer quadrotors, the reason for performance differences could be from the reduction in the size of the majority class. Not only is there a concern with class distribution when performing our feature selection technique but also with runtime and efficiency in general.

The first part of feature selection using this data involves converting ulog files. Converting raw ulog files to usable Pandas dataframes takes the most time since these files can be quite large.
Depending on the feature subset used, we may need to parse through more ulogs if they contain the feature of interest.
For example, some feature subsets lead to over 20,000 ulogs, whereas the feature subset that we are currently using has around 13000 ulogs.
Based on the approximate time it takes to convert an ulog, these additional 7000 ulogs can take several hours.
However, by pickling the data and making it open source, we can reduce this inconvenience in future research.

\subsubsection{General Classification Observations}
As expected, our model did not perform as well with hexarotor instances as it did with quadrotors and fixed-wing UAVs.
The first reason is likely because of the class imbalance.
However, the model can still predict fixed-wing instances relatively well even if there were about the same fixed-wing instances as hexarotors.
Since hexarotors are quite similar to quadrotors in terms of their hardware and flight behaviors, it makes sense to have misclassification between the two classes. 

Additionally, we can perform some analysis on the instances that were misclassified. Our initial assumption that hexarotors would be often classified as quadrotors was correct. A confusion matrix (\cref{tab:confMat}) shows  for the combined fold predictions from trial 1 (see \cref{tab:TimestampSampling} for reference). 214 out of 347 hexarotors were predicted as quadrotors, which is approximately 62\% of the instances compared to the 15 that were predicted as fixed-wing UAVs. A majority of hexarotors are classified as quadrotors; however, our algorithm is still capable of predicting 118 hexarotors correctly, or approximately 34\% of the instances. These results are promising since the difference between quadrotors and hexarotors is relatively small. Similarly but to a lesser degree, 124 out of 412 fixed-wing UAVs or approximately 30\% of the instances were predicted as quadrotors compared to the 10 that were predicted as hexarotors.

\subsubsection{Sampling}
After testing different timestamp sampling techniques, it is evident that equal-width average sampling led to the best classification performance under all tested configurations, as shown in Table \ref{tab:TimestampSampling}.
The macro f-score across the three classes and the individual precision and recall scores indicate that this method could more accurately distinguish between quadrotors and hexarotors.
This finding becomes apparent when comparing average sampling with windowed sampling.
Both techniques had similar results for the quadrotor class, but average sampling yielded a much greater f-score for the hexarotor class (42.15\% in the top performer of average sampling vs.\ less than 10\% across all configurations of windowed sampling).
Similarly, fixed-wing UAVs had higher scores when average sampling was applied (73.15\% in the top performer of average sampling vs.\ less than 66\% across all configurations of windowed sampling).

One potential reason that average sampling gave the best results is because it utilizes the most data out of the three approaches.
Fixed window average leaves the data between the end of the current window and the beginning of the next window out of the instance. This data could be informative pieces of the flight but is removed depending on the window size. With average sampling, we ensure that all data is kept and summarized accordingly.

In terms of speed, each method is quite similar and does not take much time.
As described in our methodology and shown in diagrams, the difference among the approaches is subtle.
Each approach requires the same calculation of finding the beginnings and ends of the intervals given the minimum and maximum timestamps of the instances.
Then we sample an average over a range of values or a single point accordingly, which is a negligible difference.

\subsubsection{Handling Class Imbalances}

In our experiments with handling class imbalance, we found that these techniques led to tradeoffs in performance for the different types of UAVs, as shown in Table \ref{tab:classInbalance}.
In general, the f-scores of the quadrotor class across all techniques and configurations were consistent when compared against the results with no class imbalance changes.
However, one trend we see is that the recall of the fixed-wings and hexarotor classes increases at the cost of precision.
Essentially, we predict these classes more frequently when we augment, oversample, or undersample.
These predictions are sometimes false positives, but the chance of predicting actual fixed-wings and hexarotors increases.
For example, in Table \ref{tab:percisionRecallTradeoff}, we compare against our top performer from the timestamp sampling experiments.
 We can achieve at most a 48.73\% increase in recall (using cluster centroid undersampling with a 75\% decrease in the majority class for hexarotors.
However, using this undersampling approach decreases precision by 35.97\%.

These results show that undersampling, oversampling, and data augmentation are somewhat naive approaches in our setting to gain performance when adjusting for class imbalances.
Sampling techniques either drop out instances or generate interpolated synthetic instances. Data augmentation similarly generates synthetic instances through transformation. Interpolation and transformations may not be appropriate for complex forms of high dimension data. An alternative that addresses this limitation and helps with class imbalance problem is using simulated data. Simulated data ensures that we are generating more realistic instances because of the modelling capabilities of simulation software.

Still, there are domains where simulated data is not possible and the tested techniques may be useful. 
If one decides to use any of these methods, implementation is simplified through the Tsaug and Imblearn libraries.
However, slight modifications to the shape of the data are required. For instance, Imblearn expects more traditional tabular data.
Speed only becomes a factor when using certain ML-based sampling procedures.
For example, depending on the machine used and the amount of data, cluster-centroid undersampling will take several minutes more than the other techniques listed.

\section{Limitations}
\label{sec:limitations}

As mentioned previously, the \textbf{feature selection} process has significant room for improvement.
The first part of this process is to determine candidate features.
Determining these candidate features can be done both manually and analytically. 
If done manually, more domain knowledge and research are required to determine which features can contribute to predictive ability.
A more analytical approach to prune features involves checking the variances of features across time.
For example, some features are more related to the flight's status than its motion. Therefore, the second part of this process involves either dimension reduction or a more brute-force feature selection process.
If the additional emphasis is placed on feature selection, then we can remove some of the barriers when working with this data in the future.

There are various other \textbf{timestamp sampling techniques} that might be applicable.
As mentioned in the methodology section, using equal spacing is unnecessary, and quantile-based approaches may be viable \cite{disc}.
\cite{disc} took sampling further by comparing local, global, and hybrid binning.
Our sampling is done locally, meaning each instance is sampled independently of the others.
However, the time series could be rescaled and binned with one global or hybrid approach of local and global binning.
Aside from these sampling techniques, which we will refer to as binning transformations, there are also continuous transformations that employ log, and power transforms \cite{disc} to normalize the data. 

There are a few techniques that can help with addressing the \textbf{class imbalance}.
The first involves simulated data; if we have the waypoints of a quadrotor flight, we can use a UAV simulator to fly the same flight but with the other two UAV types.
The second approach is use addressing the problem from the training side.
There are cost functions that take into account class imbalances.
For example, \cite{cost} uses a learnable parameter within the cost function that penalizes the misclassification of the minority class. 
Lastly, our loss function can handle class imbalances through sample weighting.
The algorithm can use a class-wise re-weighting scheme across any loss function, whether our cross-entropy function or another loss function approach. 
The idea is to weigh the loss computed for different samples differently based on whether they belong to the majority or minority classes.

\section{Conclusion}
\label{sec:conclusion}

In this paper, we present first of its kind work on detecting if a UAV is quadrotor, hexarotor, or fixed-wing.
We use an LSTM model with a single hidden layer to classify the type of UAV and found that our model does a good job detecting quadrotor and fixed-wind but struggles with hexarotor.
We explore how different sample methods, average and fixed window average sampling, and parameters affect precision, recall, and the f-score.
Lastly, we test several oversampling and data augmentation methods to help handle the class imbalance in the data.
In the future, we plan to tackle the limitation presented and continue work on classifying UAV behavior.



%



\section*{Acknowledgment}
This research was supported by an appointment to the Intelligence Community Postdoctoral Research Fellowship Program at Northeastern University administered by Oak Ridge Institute for Science and Education (ORISE) through an interagency agreement between the U.S.\ Department of Energy and the Office of the Director of National Intelligence (ODNI).

\ifCLASSOPTIONcaptionsoff
  \newpage
\fi



\bibliographystyle{IEEEtran}
\bibliography{references}

\end{document}